\documentclass[letterpaper, 10 pt, conference]{ieeeconf}  

\IEEEoverridecommandlockouts                              

\usepackage{graphics} 
\usepackage{epsfig} 
\usepackage{mathptmx} 
\usepackage{times} 
\usepackage{amsmath} 
\usepackage{amssymb}  
\usepackage[dvipsnames]{xcolor}
\usepackage{algorithm}
\usepackage{algpseudocode}
\usepackage{blindtext}
\usepackage{makecell}
\usepackage{mdframed}

\usepackage[noadjust]{cite}

\usepackage{soul}
\usepackage{tcolorbox}
\usepackage{ragged2e}
\usepackage{tabularx}
\usepackage{hyperref}

\newtcolorbox{grayblock}{
  colback=gray!5,
  colframe=black,
  boxrule=0.5pt,
  left=3pt,
  right=0pt,
  top=0pt,
  bottom=0pt,
  boxsep=5pt,
  arc=3pt,
  outer arc=3pt,
}
\definecolor{WildStrawberry}{HTML}{FF5132}
\definecolor{Salmon}{HTML}{FA9123}
\definecolor{Orange}{HTML}{A68200}
\definecolor{LimeGreen}{HTML}{8F9E44}
\definecolor{Turquoise}{HTML}{546774}
\definecolor{Orchid}{HTML}{700F7A}

\title{\LARGE \bf
MORPHeus: a Multimodal One-armed Robot-assisted Peeling System with Human Users In-the-loop
}

\author{Ruolin Ye, Yifei Hu, Yuhan (Anjelica) Bian, Luke Kulm, and Tapomayukh Bhattacharjee
\thanks{
        This work was partly funded by National Science Foundation IIS \#2132846, CAREER \#2238792, and DARPA under Contract HR001120C0107. We thank Daniel Stabile and Mavis Lee for their help with the data collection portal. Videos and supplementary materials are available at our website~\cite{morpheus2023}.
}
\thanks{All the authors are affiliated with Cornell University, Ithaca, NY, USA,
        {\tt\small \{ry273, yh464, yb265, lbk73, tb557\}@cornell.edu}}
        }

\renewcommand{\baselinestretch}{0.97}
\begin{document}

\maketitle
\thispagestyle{empty}
\pagestyle{empty}

\begin{abstract}
Meal preparation is an important instrumental activity of daily living~(IADL). While existing research has explored robotic assistance in meal preparation tasks such as cutting and cooking, the crucial task of peeling has received less attention. Robot-assisted peeling, conventionally a bimanual task, is challenging to deploy in the homes of care recipients using two wheelchair-mounted robot arms due to ergonomic and transferring challenges. 
This paper introduces a robot-assisted peeling system utilizing a single robotic arm and an assistive cutting board, inspired by the way individuals with one functional hand prepare meals. Our system incorporates a multimodal active perception module to determine whether an area on the food is peeled, a human-in-the-loop long-horizon planner to perform task planning while catering to a user's preference for peeling coverage, and a compliant controller to peel the food items. We demonstrate the system on 12 food items representing the extremes of different shapes, sizes, skin thickness, surface textures, skin vs flesh colors, and deformability. Check out the MORPHeus project at \url{https://emprise.cs.cornell.edu/morpheus/}.

\end{abstract}

\section{Introduction}

According to a recent report~\cite{united2014americans}, up to $24$ million people aged $18$ years or older need assistance with activities of daily living such as feeding, as well as instrumental activities of daily living such as meal preparation (peeling, cutting, cooking, etc.). 
While there is a significant amount of research involving feeding~\cite{onlinefeeding, transferfeeding, feng2019robotassisted, jenamani2024robot, RCareWorld, ondras2022human, xu2022evaluating, bhattacharjee2020autonomy, feng2019robot}, cutting~\cite{xu2023roboninja, heiden2021disect, Mu2019RoboticCM, fu2023demonstrating}, and cooking~\cite{cooking, Yi2022AnOT, Takata2022EfficientTP}, peeling is relatively underexplored. In this paper, we focus on robot-assisted peeling with a single-arm setup, 
to address the needs of people with mobility limitations who use wheelchair-mounted robot arms. 

\begin{figure}[t!] 
  \centering
  \includegraphics[width=0.9\linewidth]{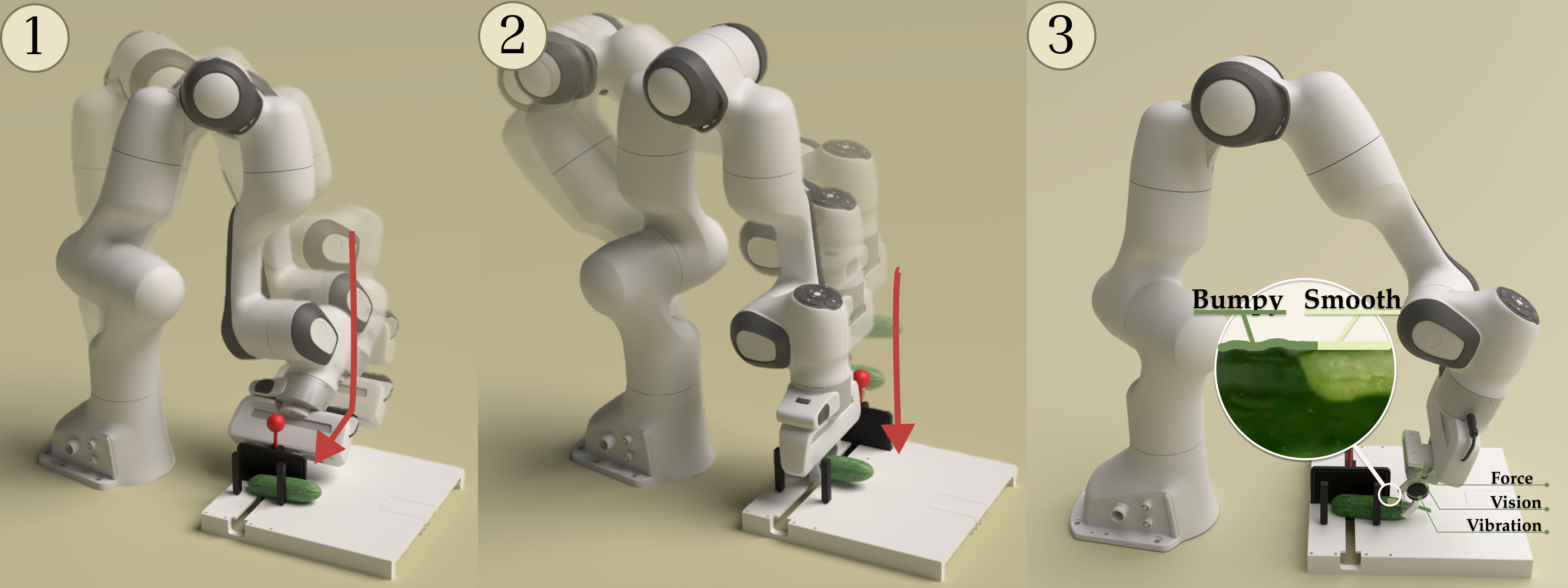} 
  \vspace{-3mm}

  \caption{Concept of robot-assisted peeling with an adaptive cutting board: (1, 2) The robot uses fixtures on the board to secure the food
 and interacts with the food, resulting in long-horizon manipulation. (3) The robot uses a peeler instrumented with multimodal sensors to identify whether an area is peeled. }
  \label{fig:teaser}
  \vspace{-6mm}
\end{figure}

Conventionally, peeling is considered a bimanual task in which one arm stabilizes the food item while the other arm executes the peeling action~\cite{peel2021}. Most real-world robotic caregiving systems~\cite{wheelchair} use one robot arm mounted on the wheelchair
because mounting two robot arms on wheelchairs is practically challenging due to ergonomic and transfer challenges for care recipients~\cite{kinova_jaco_2024}. 
We use an assistive cutting board inspired by how individuals with one functional hand prepare meals. An assistive cutting board~\cite{etac-cutting-board-review}, also known as an adaptive cutting board, is a board with multiple fixtures that can secure a food item (shown in Fig~\ref{fig:teaser}), so the user can peel with one arm. In addition, when users peel, they might have different preferences.
For example, a user might want a half-peeled cucumber if they want to prepare cucumber bites with some texture provided by the skin.
We want to ensure that their preferences about peeling coverage (fully-peeled, half-peeled, top-peeled, etc.) are taken into account. 

Robot-assisted peeling presents numerous challenges due to the varying physical properties of food items, such as their shape, size, deformability, skin thickness, and skin-to-flesh color contrast. 
Using a single robot arm for peeling tasks with adaptive cutting boards introduces complex issues. The robot needs to reason how to use the complicated fixtures on the cutting board. Also, it needs to have a control policy that can peel a wide range of food items.

Our key insight is that multimodal perception with human-in-the-loop long-horizon planning and compliant control can enable a one-armed robot to perform the peeling task for a wide variety of food items while taking users' preferences into consideration. Using this insight, we develop a system that includes: (1) a multimodal active perception module using visual, force, and vibration sensing modalities to estimate the peeling state of a food item during task execution; (2) a Large Language Model (LLM)-based interface to convert a user's natural language commands to long-horizon plans with a human user in-the-loop; (3) a compliant controller for executing the peeling motion and adapting to a wide variety of food items with varying physical properties.
Check out the MORPHeus system at \url{https://emprise.cs.cornell.edu/morpheus/}.

\begin{figure*}[ht!] 
  \centering
  \includegraphics[width=0.9\textwidth]{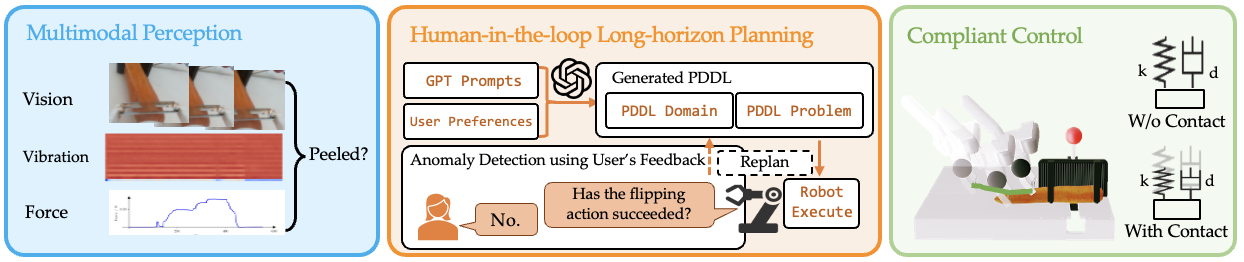} 
  \vspace{-2mm}

  \caption{Overview of the building blocks of MORPHeus: Multimodal perception (Sec.~\ref{Sec:multimodal}) module uses visual, vibration, and force inputs to determine whether the food is peeled, Human-in-the-loop Long-horizon Planning (Sec.~\ref{Planning:Integrating}) enables the robot to perform task planning while catering to a user's preference and perform anomaly detection using user's feedback, and Compliant Control (Sec.~\ref{Sec:control}) to execute peeling actions for various food items.}
  \label{Fig:Overview}
    \vspace{-3mm}
\end{figure*}

We summarize our contributions as follows:
\begin{itemize}
    \item A one-armed robot-assisted peeling system with compliant control for peeling food items with a wide variety of physical properties and various assistive cutting boards.
    \item A multimodal active perception model incorporating visual, force, and vibration inputs for peeling.
    \item A natural language interface for generating long-horizon plans that cater to user preferences with human users in-the-loop to provide feedback for peeling.
    \item A dataset of human peeling actions on a wide variety of food items and an open-hardware design of a peeler with sensors for visual, force, and vibration modalities.
\end{itemize}

\section{Related Work}

\subsection{Meal Preparation with Robots}\label{RelatedWork:FoodPreparation}
Meal preparation can be broken down into multiple substeps, including cutting~\cite{xu2023roboninja, heiden2021disect
}, peeling~\cite{peel2021}, and cooking~\cite{cooking}. Xu et al.~\cite{xu2023roboninja} developed a cutting system to cut food with solid cores, such as avocado. Liu et al.~\cite{cooking} proposed a model that learned stir-fry motions from human demonstrations. These works either assume that the food items have already been peeled or ignore the peeling problem. There is also research on peeling bananas with a dexterous hand~\cite{peelbanana}, peeling grapes with a surgical robot~\cite{davincigrape}, and peeling a grapefruit with a knife~\cite{surfing}.
Watanabe et al.~\cite{peel2013} and Dong et al.~\cite{peel2021} used peelers for peeling, similar to our setup. However, they both used a dual-arm setup: one arm stabilizes the food items, and the other peels. Watanabe et al.~\cite{peel2013} used 2D visual perception to determine whether the food was peeled and tested their system with a single food item. Dong et al.~\cite{peel2021} added a point cloud to obtain 3D information and expanded the scope to 5 food items. Our work uses visual, force, and vibration modalities to determine whether the food item is peeled, takes a variety of assistive cutting boards into consideration,
and performs experiments on 12 vegetables and fruits that cover a wide range of physical properties, with only one robot arm.
Additionally, our work differs from previous work in the fact that we perform the peeling task based on user preferences about the peeling coverage.

\subsection{Human-in-the-loop Long-horizon Planning}\label{RelatedWork:Planning}
Language serves as a natural interface to incorporate human preferences into long-horizon planning~\cite{zhao2023large, wu2023tidybot}. 
In recent years, there has been significant progress in large language models~\cite{devlin2019bert, CodeX, Opt, GPT3, gpt4, InstructGPT, llama, palm}.
Given their extensive pre-training on diverse text data, LLMs exhibit impressive zero-shot generalization capabilities, making them suitable for use in robot planning~\cite{zeroshotplanning, singh2022progprompt, vemprala2023chatgpt}. Therefore, we use LLM to perform long-horizon planning with human-in-the-loop, taking their preferences into consideration.
However, compared to directly generating task plans using LLMs, converting natural language to 
Planning Domain Definition Language (PDDL), and planning with a classical planner has the following advantages:
(1) the PDDL generated by LLM can be quickly scrutinized; (2) executing a synthesized program is often quicker than repeatedly querying the LLM for each new job; (3) synthesized programs can be customized for tasks of any size; (4) using PDDL circumvents the context window size limitations inherent to LLMs~\cite{silver2023generalized}. We therefore use LLM integrated with PDDL for long-horizon planning.
Previously, research has already focused on fine-tuning Large Language Models (LLMs) for the PDDL domain
~\cite{pallagani2022plansformer}, as well as on translating natural language into PDDL~\cite{collins2022structured, lin2023text2motion, xie2023translating, liu2023llmp}. These capabilities make LLMs well-suited for human-in-the-loop planning scenarios.
\section{Methodology}
MORPHeus has 3 building blocks: multimodal perception, long-horizon planning, and compliant control. The perception module informs the planner about the states of the peeler, the robot, the cutting board, and the food item. The planner performs human-in-the-loop long-horizon planning, integrating an LLM~(GPT-4) with PDDL. For control, we use a Cartesian impedance controller.
Fig.\ref{Fig:Overview} shows the system modules.
\subsection{Multimodal perception}\label{Sec:multimodal}
\subsubsection{Perception Model}
\begin{figure}[t] 
  \centering
  \vspace{-3mm}
  \includegraphics[width=0.4\textwidth]{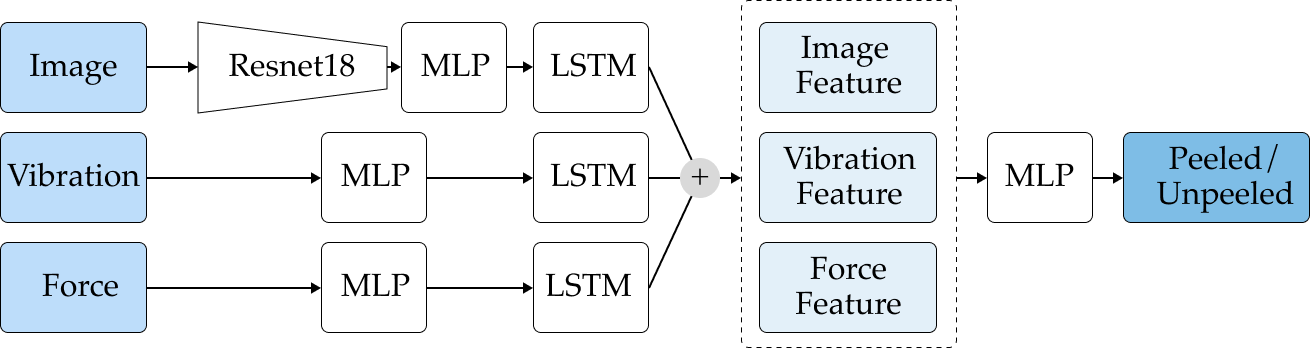} 
  \caption{\scriptsize{Model structure for multimodal perception: The network uses image, vibration, and force as input, and gives a binary output for peeled/unpeeled classification ($O$).}}
  \label{fig:model}
  \vspace{-4mm}
\end{figure}

\begin{figure*}[ht!] 
  \centering
  \includegraphics[width=0.9\textwidth]{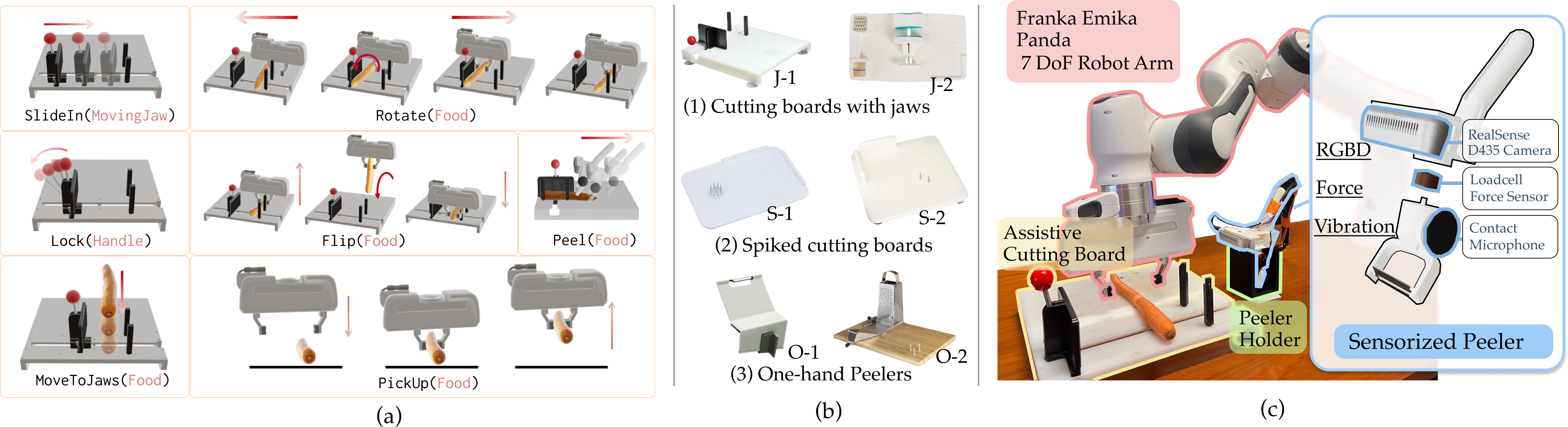} 
  \vspace{-3mm}
  \caption{(a) Actions the robot can take. For bidirectional actions including SlideIn/Out, Lock/Unlock, MoveToJaws/MoveFromJaws, we only visualize one direction here. (b) Various types of cutting boards. (c) The experiment setup includes a Franka Emika Panda 7 DoF Robot Arm, an assistive cutting board with a moving jaw, a 3D printed peeler holder, and a peeler with multimodal perception for vision, force, and vibration modalities.}
  \label{Fig:Setup}
    \vspace{-6mm}
\end{figure*}

We introduce a binary classification model that takes an image, vibration, and force sequence as inputs. 
The three data modalities go through parallel feature extraction layers, including a network for images, multi-layer perceptron networks~(MLPs), and LSTMs.
These processed data streams are then concatenated and passed through additional dense layers to produce a binary final output for peeled/unpeeled ($O$).
Given the model's output $O$ and the human-labeled ground truth, we use the Binary Cross Entropy (BCE) as the loss function. We show the structure of the model in Fig. \ref{fig:model}. Please refer to the details on our website~\cite{morpheus2023}.

\subsubsection{Multimodal Peeler}
We design a peeler equipped with multimodal sensors and release it as open hardware. Fig. \ref{Fig:Setup}(c) shows the multimodal peeler and its structural design. We use a Realsense D435 camera for visual input~\cite{realsense-d435i}, a TE FX29K0-040B-0010-L load cell sensor~\cite{loadcell} for force, and a piezo contact microphone for vibration~\cite{contact_microphone}. The force sensor is connected to an Arduino Uno Rev3 board~\cite{arduino} for readings. The contact microphone is connected to an amplifier and then to a desktop for data collection. The image signals are sampled at 30Hz, the force signals at 2.4kHz, and the vibration signals from the contact microphone at 1.6kHz. We assemble the sensors on the 3D-printed parts of the custom-designed peeler head and peeler holder, and mount a blade on the peeler similar to a standard peeler.

\begin{figure*}[t!] 
  \centering
  \includegraphics[width=0.9\textwidth]{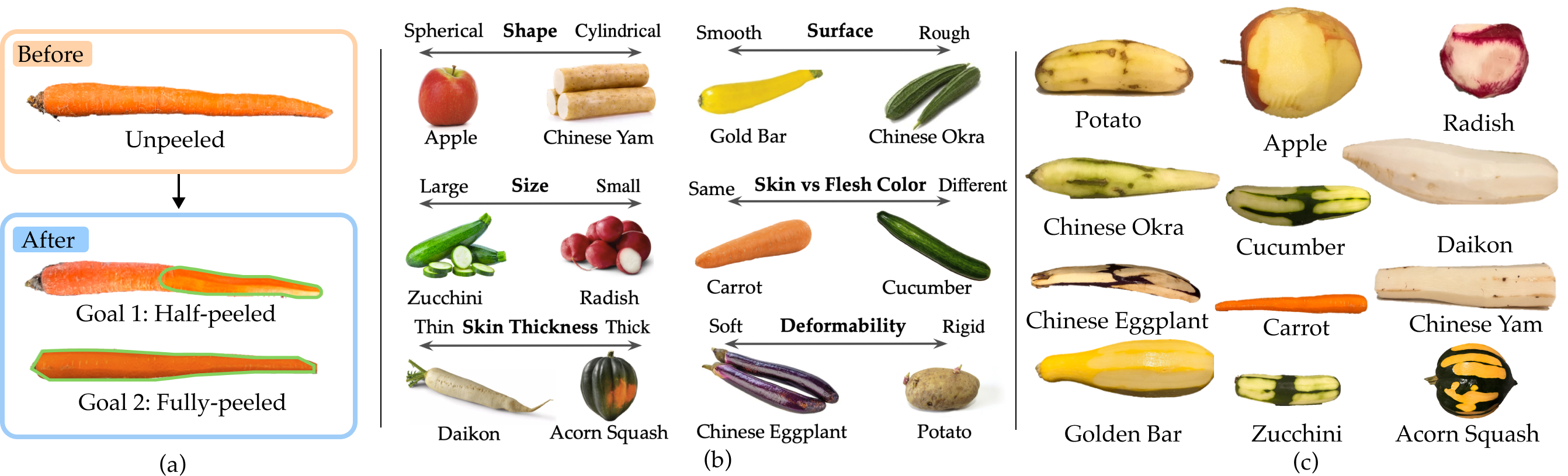} 
  \vspace{-4mm}

  \caption{(a) Users' preference: By expressing their preference for the food to be `Half-peeled' or `Fully-peeled', the users can get their food peeled according to their preferred peeling goal. (b) 6 axes corresponding to the characteristics of different food items and 12 food types selected to represent the extremes. (c) Food items peeled by MORPHeus. }
  \label{Fig:Experiments}
    \vspace{-6mm}
\end{figure*}
\vspace{-1mm}
\subsection{Human-in-the-loop Long-horizon Planning}\label{Planning:Integrating}
We use GPT-4 as a foundation model for long-horizon human-in-the-loop planning. The role of GPT-4 is to generate the files for PDDL. In planning with PDDL, there are two critical components: a PDDL domain file and a PDDL problem file. The domain file categorizes objects in the environment into different types, specifies possible actions along with their preconditions and effects, and includes a set of binary condition predicates. The problem file outlines the initial state of the environment and the desired goal state. We introduce the planning module with necessary high-level information in this section. For the details, please refer to our website~\cite{morpheus2023}.

\textbf{Prompt Specification:}
We provide GPT-4 with a structured prompt as follows.
The prompt describes 
\setulcolor{WildStrawberry} (a) \ul{the role of GPT},
\setulcolor{Salmon} (b) \ul{the robot actions}, 
\setulcolor{Orange} (c)\ul{ the environment}, 
\setulcolor{LimeGreen} (d)\ul{ the format} of the output, \setulcolor{Turquoise}(e) \ul{ example} of a customized prompt with the same structure with the generated domain PDDL, and \setulcolor{Orchid}(f)\ul{ user inputs}, such as peeling goals and feedback. We show an example of the prompt inside the gray box~\ref{prompt}-1. The prompt with these components gives GPT the context to reason for this specific peeling task, and also constraints its output syntax by giving examples of PDDL files.
Prior research, such as ~\cite{liu2023reflect, gou2023critic}, has highlighted the proficiency of Large Language Models (LLMs) in utilizing human feedback to detect and correct anomalies in generated content. In our planning approach, we implement an \textbf{anomaly detection with user's feedback} module to address potential execution errors. At critical junctures in the plan, this module will seek human feedback to detect or address potential anomalies.
After completing a rotation or a flip, the process pauses to ask the user some questions regarding whether the action has succeeded. For example, the robot may ask ``Has the rotation action succeeded?" If the user identifies an anomaly, the LLM engages the user with a series of binary questions, such as ``Is the food item between the jaws?", to update its understanding of the environment. Following this, the LLM replans to rectify the previous failure before proceeding, assume undo is possible.
\begin{grayblock}
\justifying
{\scriptsize
\label{prompt}
\texttt{
\textbf{Example Prompt~(\ref{prompt}-1): }
\textcolor{WildStrawberry}{(a) Role of GPT}
\setulcolor{WildStrawberry}
\ul{You are an excellent interpreter of human instructions for caregiving tasks...}
\textcolor{Salmon}{(b) Robot actions}
\setulcolor{Salmon}
\ul{PickUp...PutDown...}
\textcolor{Orange}{(c) Environment}
\setulcolor{Orange}
\ul{...a robot arm with a gripper...an adaptive cutting board... to peel a food item that is on the table...}
\textcolor{LimeGreen}{(d) Output format}
\setulcolor{LimeGreen}
\ul{The format of output should be a PDDL domain file...}
\textcolor{Turquoise}{(e) Examples}
\setulcolor{Turquoise}
\ul{Below is a simple example of generating a PDDL involving non-deterministic action...}
\textcolor{Orchid}{(f) User inputs}
\setulcolor{Orchid}
\ul{The user wants the potato fully peeled...}
}
}
\end{grayblock}

\textbf{Action Specification:} The peeling task involves a complex, long-horizon manipulation sequence that involves a variety of subtasks. Imagine a robot using a cutting board with a moving jaw (J-1 and J-2 in Fig. \ref{Fig:Setup}(b)) to peel a carrot: the robot first slides in the moving jaw and locks it to secure the food item. Then, it peels the food. After peeling, it unlocks the moving jaw, pulls out the fixture, and then rotates the food to proceed to the next peeling action. If the robot uses the spiked cutting board (S-1 and S-2 in Fig. \ref{Fig:Setup}(b)), it puts the carrot on the spikes, and then peels the carrot. The subtasks for peeling are different for different cutting boards.
For these subtasks, we combine possible actions across three different types of cutting boards to create an action space. The action space includes a description of the cutting boards and the accompanying actions.
This enables the LLM to mix and match action sets for planning when multiple types of cutting boards are available. We illustrate actions specific to the cutting board with jaws, as used in our experimental setting, in Fig.\ref{Fig:Setup}(a). 
For a detailed implementation of each action and the actions for other cutting boards, please refer to our website~\cite{morpheus2023}.

\subsection{Compliant Control}\label{Sec:control}
To ensure effective peeling without damaging the flesh of food items, the robot needs to exert the appropriate amount of force on the uneven food surfaces. We achieve this by using a Cartesian impedance controller that dynamically adjusts the applied force when peeling. 
Let $\mathbf{q} \in \mathbb{R}^7$ be joint positions (7 DoF robot arm), $\dot{\mathbf{q}} \in \mathbb{R}^7$ be joint velocities, $\mathbf{c}(\mathbf{q}, \dot{\mathbf{q}}) \in \mathbb{R}^7$ be Coriolis torque, and $J(\mathbf{q}) \in \mathbb{R}^{6 \times 7}$ be the Jacobian matrix. We define the translational stiffness as $k_t$ and rotational stiffness as $k_r$, with which we can construct the stiffness matrix $K$ and damping matrix $D$ as the following assume critical damping:
$$
\begin{gathered}
K=\left[\begin{array}{ll}
k_t I_3 & 0_{3 \times 3} \\
0_{3 \times 3} & k_r I_3
\end{array}\right], 
 D=\left[\begin{array}{cc}
2 \sqrt{k_t} I_3 & 0_{3 \times 3} \\
0_{3 \times 3} & 2 \sqrt{k_r} I_3
\end{array}\right],
\end{gathered}
$$ where $I_3$ is the $3 \times 3$ identity matrix. 

Let $\mathbf{x} = FK(\mathbf{q}) \in \mathbb{R}^6$ be the task space pose, where $FK(\cdot)$ is the forward kinematics module, and $\mathbf{x}_d \in \mathbb{R}^6$ be the desired task space pose. $\mathbf{f_{ext}}$ is the desired external force in the task space. With these definitions, we use the control law in Eq.~\ref{eqn1} to obtain the desired force:
\begin{equation}\label{eqn1}
    \mathbf{f_{ext}} = K(\mathbf{x}_d - FK(\mathbf{q})) + D(J\dot{\mathbf{q}})
\end{equation}

\section{System Component Evaluations}
We evaluate each of the system components separately with a combination of simulations and real-world experiments.

\subsection{Multimodal Perception}
\subsubsection{Experiment setup} We evaluate the multimodal perception module using the setup in Fig. \ref{Fig:Setup}(c). We use a Franka Emika Panda 7DoF robot arm and put a cutting board in front of it. The cutting board is fixed to the table using suction cups. We also design a 3D-printed holder for the peeler, and fix it to the table.
\paragraph{Food selection}
To ensure food selection with a wide variety of physical properties, we consider the following characteristics: (1) shape, (2) surface texture, (3) size, (4) contrast of color between skin and flesh, (5) skin thickness, and (6) deformability. 
For each of them, we select two food items that represent the extremes, yielding a total of 12 distinct food items with significantly varying characteristics. We show the 12 food items and their different characteristics in Fig. \ref{Fig:Experiments}(b).
\paragraph{Data Collection Procedure}
We collect the data for multimodal perception by performing a set of haptic exploratory actions where the robot uses a set of specific \textit{action primitives} to figure out the haptic properties of the food items. The \textit{action primitives} include sliding over the unpeeled surface, peeling over the unpeeled surface, sliding over the peeled surface, and peeling over the peeled surface. For sliding, we turn the peeler and use the back of the blade to slide over the surface of the food item. We identify these actions as potential ways a robot can use to interact with the food and identify whether it is peeled or not. The purpose of this experiment is to find the most suitable actions through the haptic exploratory procedure.
We collect 100 samples, where 1 sample contains 4 action primitives. There are a total of $100\times4\times12 = 4800$ trials across all of the food items.

We evaluate the perception module by simulating a real-world setup with the dataset. We stream the data for each food item 10 times and calculate the classification success rate. We set the input length of the LSTM to 10 considering the speed of the robot. For initialization, we repeat the first frame for 10 times. As new frames come in, we append them to the input buffer. 

Based on the experiment results in Table~\ref{Real Experiment Data}, we realize that the sliding motion gives a better outcome for classifying peeled/unpeeled areas. While the visual modalities remain the same, the key difference between sliding and peeling lies in the source of vibration and force. When peeling, the front of the blade makes contact with the food item, and the blade can move freely. While sliding, the blade reaches its kinematic limit and becomes rigidly locked. As a result, less noise is generated from sliding, allowing for subtle vibrations to be more easily detected. This motivates us to use the sliding motion to tell whether the food item is peeled.

\subsection{Human-in-the-loop Long-horizon Planning}
To evaluate the human-in-the-loop long-horizon planning module, we utilize GPT-4's Chat Completion API, a context-aware text-completion API. We refer to this as `simulation'.

\indent 
 The planner's goal in this simulated experiment is to find a sequence of actions based on the domain PDDL files to transition an initial state into a goal state described in the problem PDDL files. Since the planner uses both the domain and problem PDDL files, errors in either one of them might lead to errors.
 We evaluate the following cases: (1) GPT-generated PDDL domain files~(GPT-Domain) with PDDL problem files written by human experts~(Fixed-Problem), (2) GPT-generated PDDL problem files~(GPT-Problem) with PDDL domain files written by human experts~(Fixed-Domain), and (3) GPT-generated PDDL domain and problem files together~(GPT-Domain \& GPT-Problem).
 
\indent We identify three common starting states for food items in the problem PDDL file: unpeeled, half-peeled, and top-peeled (where only the part facing the peeler is peeled). The peeling goal is fully peeled. We generate 35 domain PDDL files and 35 problem PDDL files to ensure the simulation is repeated an adequate amount of times, under each starting state using the previously proposed natural language prompt template. We use a planner with non-deterministic states, PRP (Planner for Relevant Policies)~\cite{prp2}, to find the planning policies. We run PRP for 100 rounds of simulations for each generated file, with the three different start states. 
During these simulations, the PRP uses the planning policy to generate plans across various simulated scenarios.\\
\indent In our PDDL formulation, actions like \texttt{Peel}, \texttt{Flip}, and \texttt{Rotate} might fail in the real-world setup, therefore we treat them as non-deterministic actions. For these actions, the planner selects the subsequent states uniformly at random. For PDDL generation, we use the cutting board (J-1 and J-2). We refer to trials that can run directly without any human intervention to fix syntax errors as No Attention Given~(NAG). 

\begin{table}[t!]
\caption{\small Evaluating the Multimodal Perception and Cartesian Impedance Control Components}\label{Real Experiment Data}
\vspace{-1mm}
\centering
{
\begin{tabular}{ ccc }

\begin{tabular}{l|}
\hline
   \\
\multicolumn{1}{c|}{Food name}   \\ \hline\hline
Cucumber    \\       
Carrot      \\       
Chinese Yam      \\  
Chinese Okra     \\  
Chinese Eggplant \\  
Potato      \\ 
Zucchini    \\       
Gold Bar    \\       
Daikon      \\       
Apple       \\       
Radish      \\       
Acorn Squash\\ \hline      
\end{tabular}
\centering
\begin{tabular}{|cc|}
\hline
\multicolumn{2}{|c|}{Perception}\\
\hline
Peel & Slide \\ \hline\hline
7/10        &   9/10     \\       
5/10       &   8/10      \\       
9/10       &   9/10       \\       
6/10      &   7/10      \\       
7/10      &   10/10     \\       
6/10     &   10/10    \\ 
8/10       &   9/10       \\       
5/10     &   7/10     \\       
6/10     &   7/10      \\       
7/10    &   8/10        \\       
6/10       &    4/10       \\       
4/10     &   5/10      \\ \hline      
\end{tabular}
\centering
\begin{tabular}{|ccc}
\hline
\multicolumn{3}{|c}{Control}\\
\hline
$l$     &   $m$     & $h$  \\ \hline \hline
2/10    &   \textbf{10}/10     &    10/10     \\
1/10      &  9/10      &  \textbf{10}/10      \\
6/10     &   \textbf{10}/10     &   10/10   \\
2/10     &  5/10      &    \textbf{10}/10     \\
0/10     &   7/10     &    \textbf{10}/10     \\
0/10     &  2/10      &    \textbf{10}/10     \\
3/10     &   \textbf{10}/10     &    10/10  \\
2/10     &   8/10     &    \textbf{10}/10     \\
3/10      &  9/10      &   \textbf{10}/10      \\
2/10     &   \textbf{9}/10     &   7/10      \\
\textbf{3}/10     &   2/10     &   0/10      \\
1/10     &   0/10     &    \textbf{5}/10     \\   \hline
\end{tabular}
\end{tabular}}\\
\scriptsize{
\begin{flushleft}
Peel \& Slide columns: Success rates of the perception module;  $l$, $m$, $h$ columns: success rates for low, medium, and high stiffness impedance controllers respectively.\label{fn2}
\end{flushleft}}
\vspace{-8mm}
\end{table}

We show the results in Table~\ref{PDDL Data}. The results suggest the GPT-generated domain and problem PDDL files can run robustly without any syntax errors, and are able to perform the long-horizon planning task for peeling using the cutting board with the moving jaw~(J-1, J-2). For the other cutting boards~(S-1, S-2, O-1, and O-2), we inform GPT of the cutting board action correspondence, allowing it to potentially reason how to use other cutting boards.

\subsection{Compliant Control}
To evaluate the compliant module, we use the same setup in the real world as the multimodal perception module. We evaluate how the peeling action works on the food items. We let the robot perform the peeling action with the specified end position, and manually rotate the food item after each peeling action. We repeat this 10 times for each food item. The trial is considered successful if the skin is separated from the flesh completely. We tune 3 controllers for the food items, namely low-stiffness compliance ($l$) with $k_t=120$ and $k_r=0$, medium-stiffness ($m$) compliance with $k_t=150$ and $k_r=5$, high-stiffness compliance ($h$) with $k_t=180$ and $k_r=10$
We show the results in Table~\ref{Real Experiment Data}. The results suggest that for most of the food items, the stiffness of the controllers matters. While high-stiffness works for most food items, medium-stiffness or low-stiffness controllers work better for softer or spherical food items. We select the controller with the best performance in terms of peeling success rate for each of the food items. If there are two controllers with the same performance, we select the more compliant one to avoid potential damage to the food items. 
\begin{figure*}[t!] 
  \centering
  \includegraphics[width=0.9\textwidth]{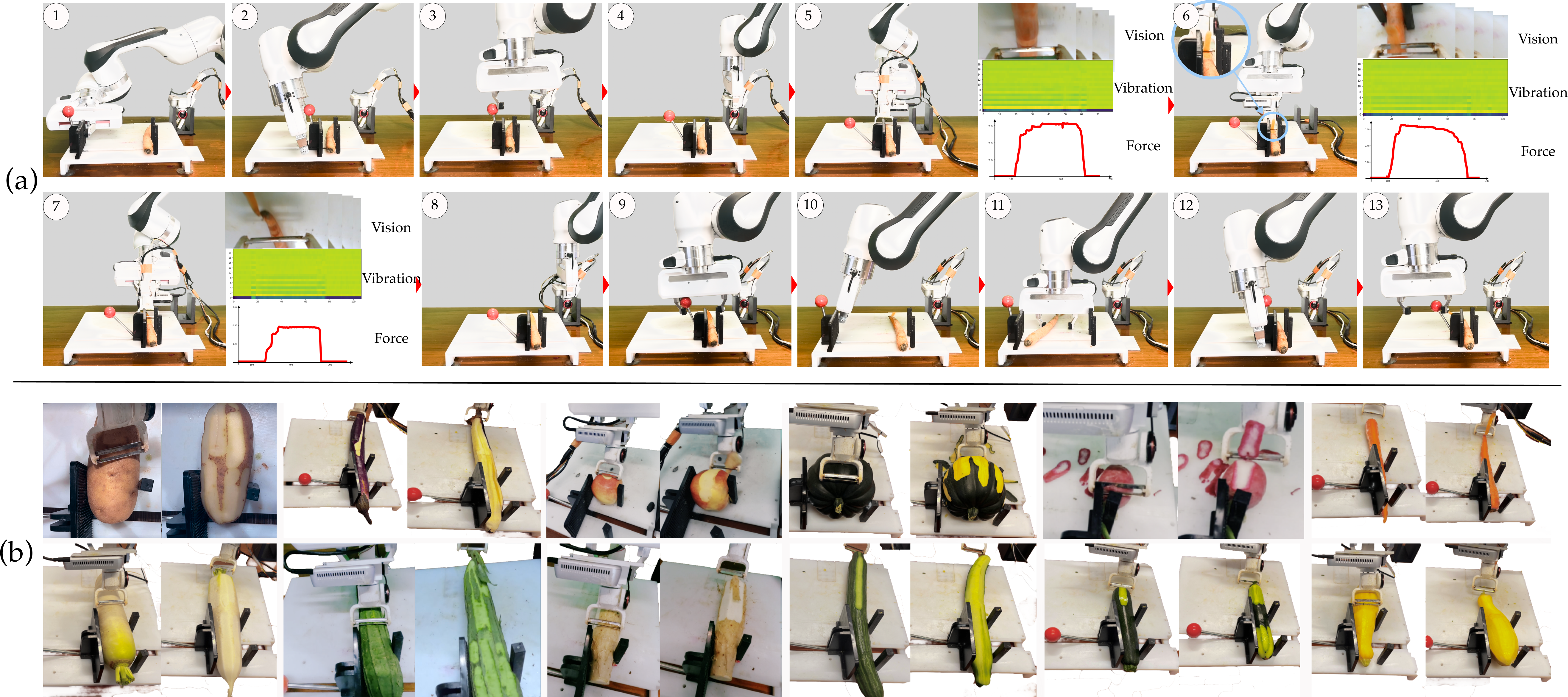} 
      \vspace{-2mm}
  \caption{(a) An example robot execution sequence for carrot peeling. (b) 12 food items at the beginning stage of peeling and at the final stage of peeling.}
  \label{Fig:RealRobot}
    \vspace{-4mm}
\end{figure*}

\begin{table}[t!]
\caption{Evaluating the planning component}
\vspace{-1mm}
\label{PDDL Data}
\centering
\resizebox{\columnwidth}{!}{
\begin{tabular}{ l  c  c  c  c  c }
\hline 
\makecell{Start\\ states}& \makecell{{NAG}$^1$}&  \makecell{Avg. \\Action} &\makecell{Max.\\ Action} & \makecell{Min. \\Action} \\ \hline \hline
\multicolumn{1}{l}{} & \multicolumn{1}{c}{} &\multicolumn{2}{c}{{GPT-Domain $\&$ Fixed-Problem}$^2$ } & \multicolumn{1}{c}{} \\ \hline 
Unpeeled       &             28/35                &       30.96$\pm$15.23   
               &            33.62$\pm$9.81   &28.70$\pm$14.31  \\
Top Peeled   & 28/35              &\textbf{24.54$\pm$12.45}   &26.39$\pm$7.30   &28.36$\pm$11.32 \\
Half Peeled           &28/35  &\textbf{22.58$\pm$8.19}   &26.44$\pm$7.21   &20.97$\pm$8.64  \\ \hline
\multicolumn{1}{l}{ } & \multicolumn{1}{c}{ } &\multicolumn{2}{c}{{Fixed-Domain \& GPT-Problem}$^3$ } & \multicolumn{1}{c}{ }  \\ \hline 
Unpeeled           & \textbf{31/35} &30.84$\pm$14.56   &33.21$\pm$11.41   &28.39$\pm$13.10  \\
Top Peeled          & \textbf{31/35}  &28.47$\pm$13.22   &29.57$\pm$12.96   &24.73$\pm$13.29 \\
Half Peeled           & \textbf{31/35}  &24.12$\pm$8.48   &28.65$\pm$11.30   &20.78$\pm$9.06  \\ \hline
\multicolumn{1}{l}{ } & \multicolumn{1}{c}{ } &\multicolumn{2}{c}{{GPT-Domain \& GPT-Problem}$^4$} & \multicolumn{1}{c}{ } \\ \hline
Unpeeled          & 27/35   &\textbf{29.13$\pm$11.42}   &32.77$\pm$9.08   &28.47$\pm$16.94  \\
Top Peeled           & 27/35 &26.57$\pm$10.20   &28.04$\pm$13.26   &22.98$\pm$13.83 \\
Half Peeled           & 27/35  &24.31$\pm$10.04   &28.79$\pm$8.92  &22.57$\pm$12.96  \\ \hline

\end{tabular}}
\begin{flushleft}
1\scriptsize{{\textbf{NAG (No Attention Given)}: Simulation runs successfully without human intervention for syntax errors.}}
2 \scriptsize{{\textbf{GPT-Domain \& Fixed-Problem}: GPT-generated domain PDDL, human-written problem PDDL.}}
3 \scriptsize{{\textbf{Fixed-Domain \& GPT-Problem}: Human-written domain PDDL, GPT-generated problem PDDL.}}
4 \scriptsize{\footnotemark{\textbf{GPT-Domain \& GPT-Problem}: GPT-generated domain PDDL, GPT-generated problem PDDL.}}
\end{flushleft}
\vspace{-6mm}
\end{table}

\section{Full Autonomous System (MORPHeus) Evaluation}
After evaluating the components individually, we evaluate the overall performance of MORPHeus by combining them together and autonomously running the whole system. We evaluate it based on the following criteria: 
\begin{itemize}
    \item Time taken to finish the task ($t$): The current State-of-the-art~(SOTA) method~\cite{peel2021} takes approximately 60 minutes to peel a single food item using a dual-arm setup. We record the time taken to peel the food item in minutes and divide it by the time of the SOTA method. $t = t_a/t_s$, where $t_a$ stands for actual time, and $t_s$ is the SOTA time~\cite{peel2021} in minutes.
    \item Peeled area ratio ($p$): We define $p$ as $p = a_p/a_a$ where $a_p$ represents the area of the peeled region, and $a_a$ represents the area of the entire region of the food surface. For simplicity, we take 2 photos, one for the front and the other for the back of the food item, ensuring as much area as possible is visible in the image. We then calculate $a_p, a_a$ in the pixel space by manually annotating the contour of the peeled area and the entire visible area of the food item. 
    \item Percentage of failed subtasks ($n_h$): In the long-horizon sequence, we consider a subtask as failed if it fails after trying 3 times. If this happens, the robot will ask the human caregiver to give it assistance to complete the subtask. We record $n_h$ based on the percentage of failed subtasks out of an average number of subtasks of 50.
\end{itemize}
Table~\ref{overallsystem} shows the results.
We noticed that the cylindrical food items have a higher success rate and a peeled area ratio, and take less time to finish. The spherical ones are easier to slip out of the jaws of the cutting board due to the design of the cutting board with a moving jaw we use~(J-1), resulting in a lower success rate and the peeled area ratio, also taking more time to finish.

\begin{table}[t]
\centering
\caption{\small Overall system evaluation}
\vspace{-1mm}
\label{overallsystem}
\begin{minipage}[t]{0.3\linewidth}
\centering
\begin{tabular}{l}
\hline
Food name   \\ \hline\hline
Cucumber    \\       
Carrot      \\       
Chinese Yam \\  
Chinese Okra     \\  
Chinese Eggplant \\  
Potato      \\ 
Zucchini    \\       
Gold Bar    \\       
Daikon      \\       
Apple       \\       
Radish      \\       
Acorn Squash\\ \hline      
\end{tabular}
\end{minipage}
\begin{minipage}[t]{0.4\linewidth}
\centering
\begin{tabular}{ccc}
\hline
$t$    & $p (\%)$    & $n_h (\%)$ \\ \hline \hline
0.3    &   93.4     &  6      \\
0.27   &  90.1      &   4  \\
0.33   &   89.7     &    4     \\
0.38   &   88.1     &   2      \\
0.33   &  87.6      &   6      \\
0.57   &   85.7     &  12       \\
0.3    &   84.2     &   8      \\
0.33   &  83.7      &    10    \\
0.27   &   82.7     &  24      \\
0.46   &  76.8      &   24     \\
0.58   &   53.2     &    22    \\
0.72   &   46.3     &   33    \\
\hline
\end{tabular}
\label{fig:figure2}
\end{minipage}
\begin{minipage}[h]{0.2\linewidth}
\includegraphics[width=\linewidth]{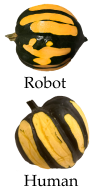}
\label{fig:figure1}
\end{minipage}\\
\scriptsize{
\begin{flushleft}
\textbf{Left:} Overall system evaluation results. $T$ column: Ratio of Time taken/SOTA time to complete the task, $p$ column: Peeled area ratio, $n_h$ column: Percentage of failed subtasks out of an average of around 50 subtasks. \textbf{Right: }Robot-peeled acorn squash compared with human-peeled acorn squash. The human uses only the dominant hand with the same peeler and cutting board as the robot for fair comparison.
\end{flushleft}}
\vspace{-6mm}
\end{table}

\section{Discussion}
\begin{figure}[t!] 
  \centering
  \includegraphics[width=0.9\linewidth]{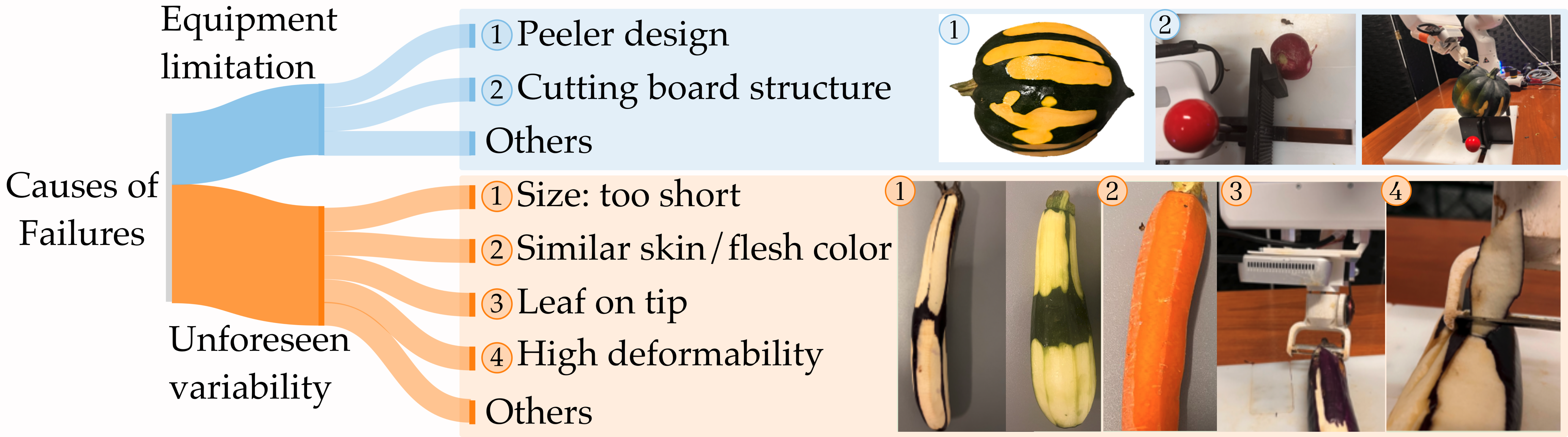} 
  \caption{Causes of failures: both equipment limitations and unforeseen variabilities may result in robot failure.}
  \label{Fig:Discussion}
    \vspace{-6mm}
\end{figure}

Based on the autonomous system evaluation, we identified the following points that can potentially help increase the success rate for future improvements, shown in Fig.~\ref{Fig:Discussion}.

\paragraph{Using suitable cutting boards and peelers based on the shape of food items~(Fig. \ref{Fig:Discussion} Blue)}
The unique groove pattern on acorn squashes complicates the peeling task when using conventional peelers. As illustrated in Table~\ref{overallsystem}, both the human and the robot are not able to reach the groves. In this setup, the human uses one arm to peel with the assistive cutting board J-1 for fair comparison.
Alternative approaches, such as using a narrower peeler, pre-cutting along the grooves, or using a knife, could be effective for a more thorough peeling. The design of the cutting board also presents challenges for handling food items that are either too large or too small. Items that are too large may slip out if their widest dimension exceeds the jaw capacity, while small or spherical items risk evading the jaw's grasp altogether. We recommend using a larger assistive cutting board to accommodate larger food items, or a spiked cutting board might also work.
The size can vary significantly even within the same category of food items. Items that are too small may not be completely peeled as a result of the cutting board's dual-peg design, which could obscure the middle section before or after flipping. Using the spiked cutting board might help address this problem.

\paragraph{Accounting for more variabilities~(Fig. \ref{Fig:Discussion} Orange)}
We identify that there are other variabilities in food items. The presence of thin skin strips may impact the robot's perception, which could be corrected through user feedback. However, in cases like carrots, where the skin and flesh colors are similar, even humans may not be able to tell the difference.
Occasionally, the eggplant peels may not completely detach, confusing the perception algorithm. Implementing a purtubation module that shakes the food and peeler could help remove the peels.
Eggplant leaves on the tip can also block the peeler. Implementing an adaptive impedance controller with online compliance updates could mitigate these issues, and we leave this open for future work.

\paragraph{Performing anomaly detection using user's feedback more intelligently}
Our experiments demonstrate that this feedback allows the robot to identify and correct execution errors by providing feedback, avoiding more substantial failures later in the plan. Limiting the feedback to yes-or-no questions has the potential to lower the cognitive workload on users with mobility limitations, compared to letting users provide detailed instructions or feedback to the robot. However, other querying types with more details, such as annotating the contour of peeled/unpeeled areas, might provide the robot with more information. For future work, a query module that can intelligently decide the query timing and type has the potential to improve success rates and lower users' cognitive workload.
\bibliographystyle{ieeetr}
\bibliography{references}
\end{document}